\documentclass{article}
\usepackage{nips14submit_e,times}
\usepackage{hyperref}
\usepackage{url}
\usepackage{amsfonts, amsmath, amssymb}
\usepackage{epsfig, subfigure, subfloat, graphicx}
\usepackage{wrapfig}

\usepackage{pgfplots}
\usepackage{tikz}
\usetikzlibrary{positioning,chains,fit,shapes,calc}
\usepackage{tkz-graph}
\usetikzlibrary{arrows}
\usetikzlibrary{decorations.text}
\usetikzlibrary{decorations.pathmorphing}

\title{Low Latency Anomaly Detection and Bayesian Network Prediction of Anomaly Likelihood}

\author{
Derek Farren$^*$ \\
\texttt{dfarren@stanford.edu} \\
\And 
Thai T. Pham\thanks{ indicates
equal contribution.} \\
\texttt{thaipham@stanford.edu} \\
\And
Marco Alban-Hidalgo \\
\texttt{marcoal@stanford.edu} 
}

\nipsfinalcopy

\begin{document}
\maketitle
\begin{abstract}
We develop a supervised machine learning model that detects anomalies in systems in real time. Our model processes unbounded streams of data into time series which then form the basis of a low-latency anomaly detection model. Moreover, we extend our preliminary goal of just anomaly detection to simultaneous anomaly prediction. We approach this very challenging problem by developing a Bayesian Network framework that captures the information about the parameters of the lagged regressors calibrated in the first part of our approach and use this structure to learn local conditional probability distributions.
\end{abstract}

\section{Introduction}

\subsection{Problem Definition}
Given a very large dataset of time series (presumably with some underlying, non-stationary time dependence between each other) up to time $t$, predict whether a time series $s$ will have an anomaly (defined formally in section \ref{sec:detecting}) at time $t+1$, moreover when the data stream arrives at time $t+1$ detect anomalies conclusively using a low latency model. 

\label{sec:Introduction}
\subsection{Overview}
The underlying motivation of this piece is the usefulness of anomaly prediction in mission critical components. A fast anomaly detection platform can by itself be extremely useful for ensuring the reliability of a system, as a result, this problem has been studied extensively by academics and industry experts. We defer to the authors in \cite{survey} for a literature survey of anomaly detection techniques.

In this work, we extend the scope of our approach and goal to tackle anomaly prediction. Most work in the literature has narrowed their focus on anomaly detection because it is in and of itself a very challenging problem. However, the benefits of an even slight temporal advantage in prediction can have huge impacts on the performance of a system. The contribution of this work can be categorized in two components:

\begin{itemize}

    \item A highly accurate, low-latency anomaly detection system(described in detail in section \ref{sec:model}) where we seek to improve on some of the concepts and techniques introduced in \cite{granger, russell}. 
    
    \item An novel approach to anomaly prediction by modeling a Bayesian Network structured based on the coefficients of the lag regressors in the anomaly detection system, coupled with Bayesian Parameter learning to model the conditional dependency structure between the time series. 
    
\end{itemize}

\section{Dataset}
\label{sec:data}
Our (non-public) data set consists of $100,000$ time series $S$, sampled every minute for the past year with some underlying non-stationary dependence between each other and human labeled anomaly events by domain experts (i.e., the owner of a time series flagged the $i^{th}$ time series $s^{i}$ at time point $j$ as an anomaly. Although the dataset has some time series source information, our anomaly detection model makes no assumptions about the inter-independence of the time series to allow us to solve the more general problem of having no apriori knowledge. 

\begin{figure}[h]
\begin{center}
\includegraphics[scale=.36]{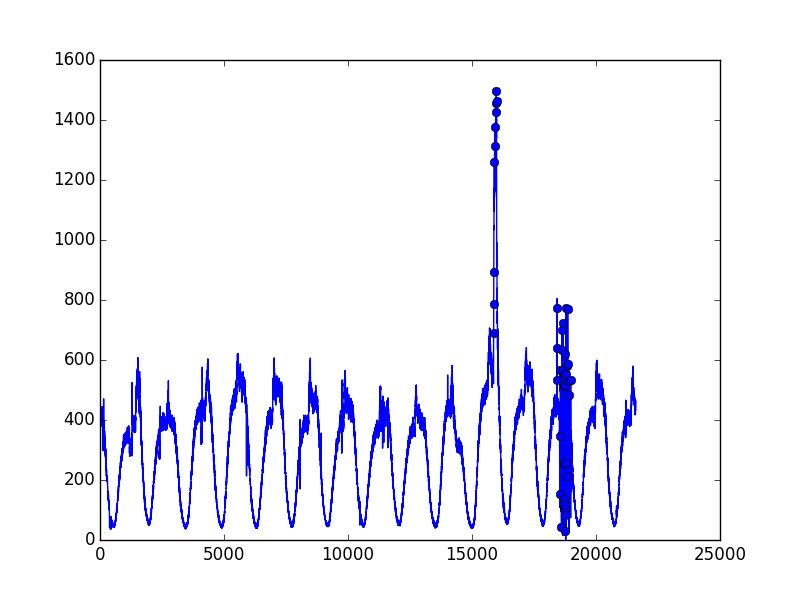} 
\caption{Subsample of A Time Series with Labeled Anomalies.}
\label{fig:sample}
\end{center}
\end{figure}

To illustrate the difficulty of the problem  we are trying to solve, in Figure \ref{fig:sample} we show a subsample of 300 points of one of the time series with labels anomalies denoted by red dots. Clearly the problem involves latent relationships between time series; observing one time series in isolation is not enough, even for a human, to determine whether a time point is anomalous. 

\begin{figure}[h]
\begin{center}
\includegraphics[scale=0.36]{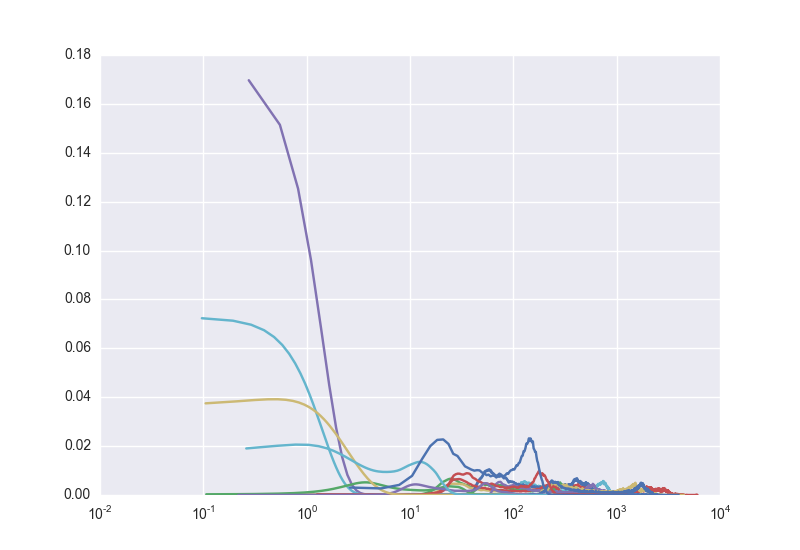}
\caption{Approximation of Probability Distribution of Selected Time Series Using Kernel Density.}
\label{fig:kde}
\end{center}
\end{figure}

Another challenge that we face is that the time series have vastly different probability distributions. To show this, we normalize the time series and approximate the probability distribution of each time series using kernel density estimation (with a Gaussian kernel and hyperparameter search on the bandwidth). Figure \ref{fig:kde} shows the estimated probability densities of the 19 most important time series. 

\subsection{Validating Dependence Assumptions}
A foundational tenet of our model is that our time series have a latent dependence structure between each other. To validate this assumption, for each pair of time series $s^{(i)}$ and $s^{(j)}$ (from a selected set of 19), we use kernel density estimation to approximate the marginal distributions of the times series, $\mathbb{P}(s^{(i)})$ and $\mathbb{P}(s^{(j)})$, as well as the joint probability distribution $\mathbb{P}(s^{(i)}, s^{(j)})$. Then we compute the mutual information between the approximate distributions:
$$ I(S_i, S_j) = \int_{s^{(i)}}\int_{s^{(j)}}p(s^{(i)}, s^{(j)}) log\Big( \frac{p(s^{(i)}, s^{(j)})}{p(s^{(i)}), p(s^{(j)})} \Big).$$

\begin{figure}[h]
\begin{center}
\includegraphics[scale=0.36]{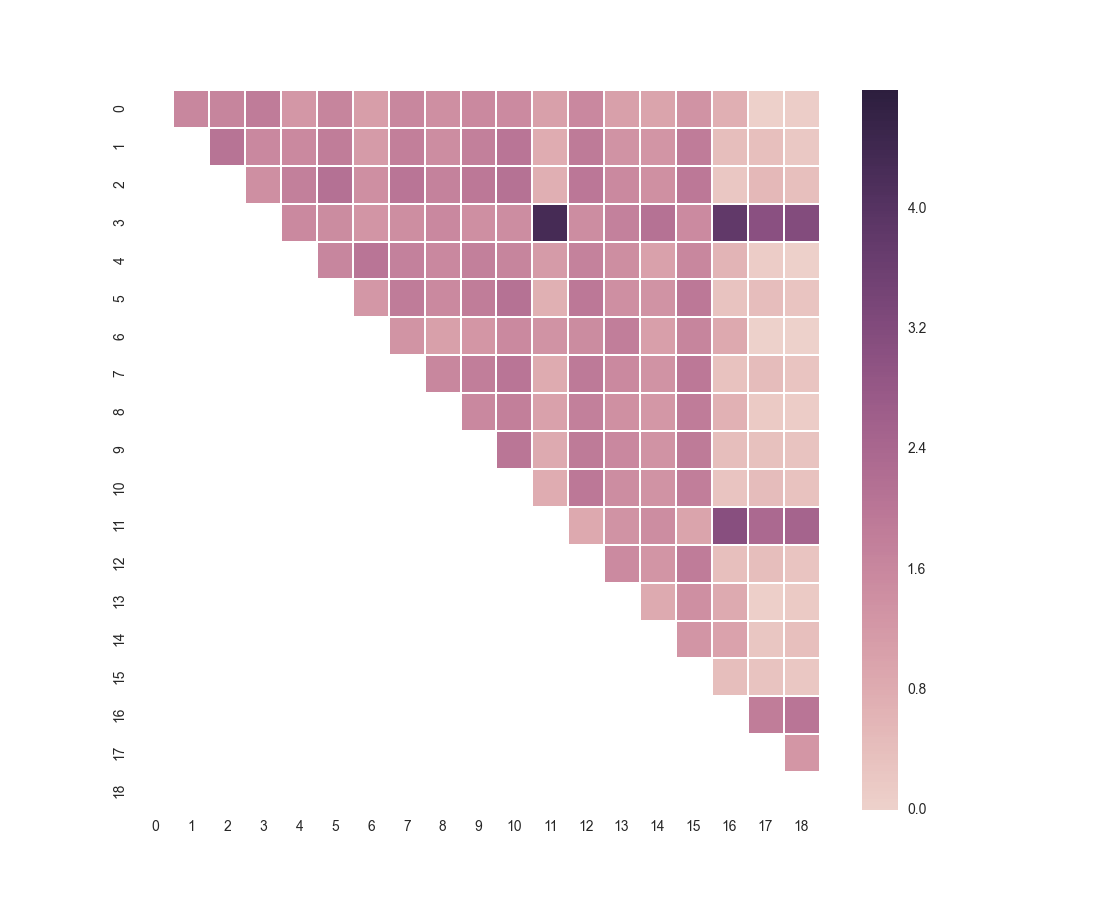}
\caption{Mutual Information Between Pairs of Time Series.}
\label{fig:MI}
\end{center}
\end{figure}

Note that mutual information is a measure of the inherent dependence expressed in the joint distribution of $X$ and $Y$ relative to the joint distribution of $X$ and $Y$ under the assumption of independence, in a concrete sense $I(X; Y) = 0$ if and only if X and Y are independent random variables. Figure \ref{fig:MI} shows a grid plot of the mutual information between 19 selected time series, the $(i,j)$ coordinate of the figure denotes $I(S_i, S_j)$. As can be seen from the figure below, the time series are highly dependent between each other, moreover, the dependence structure is not uniform across pairs. Although we could use a model like this to try to determine latent dependence structure, this method is not scalable as approximating joint probability distributions can be computationally intensive. In section \ref{sec:model} we propose a method that will capture latent dependence structures in a computationally feasible way.

\section{Approach - Detection Model}
\label{sec:model}
Combining previous techniques for anomaly detection \cite{russell, granger, survey} and time series modeling \cite{ar1, ar2, ar3, ar4}, we propose an auto regressive distributed lag model \footnote{A distributed-lag model is a dynamic model in which the effect of a regressor $x$ on $y$ occurs over time rather than all at once.} with $L_1$ regularization, that is horizontally scalable. 

We first define the following:
\begin{itemize}
    \item $n$: number of time series
    \item $s_k^{(i)}$: value of time series $i$ at timestamp $k$
    \item $w$: the window size used in the auto regression
    \item $h$: the length of the time series
\end{itemize}
Our model is
\begin{equation*}
p_{t}^{(i)}= \sum_{k=t-w}^{t-1}\sum_{j=1}^n \beta_{j,k}^{(i)} s_k^{(j)} + \epsilon_t^{(i)}
\end{equation*}
where $p_{t}^{(i)}$ is the prediction of model $i$ at time $t$, $\beta^{(i)}$ is the least squares coefficient vector, and $\epsilon_t^{(i)}$ is the model's error on time series $i$ at time $t$. In other words, we use all past information from all time series $s_k^{(j)}$ where $j \in \{1, ..., n\}$, $k \in \{t-w, ..., t-1\}$ for the prediction of the current value $p_{t}^{(i)}$ in each time series $(i)$.

Now consider concatenating all the predictions for time series $i$ (there are $h - w$ predictions). For each time series $i$, we can write the $h - w$ regression predictions as $t$ varies in the matrix form:
\begin{equation*} 
    p^{(i)} = X \beta^{(i)}  + \epsilon^{(i)},
\end{equation*}
where $p^{(i)}, \epsilon^{(i)} \in \mathbb{R}^{(h-w) \times 1}$ are the vector of predictions and error term, $X \in \mathbb{R}^{(h-w) \times wn}$ and $\beta^{(i)} \in \mathbb{R}^{wn \times 1}$. X is the matrix of concatenated lag regressors.

The model is trained using ordinary least squares method with $L_1$-regularization. In other words, for each time series $(i)$, we find the parameter vector $\beta^{(i)}$ of length $wn$ to minimize
\begin{equation}
\begin{split}
J(\beta^{(i)}) & = \sum_{t=w+1}^{h} \left[p_t^{(i)} - \sum_{k=t-w}^{t-1}\sum_{j=1}^n \beta_{j,k}^{(i)} s_k^{(j)}\right]^2 + \lambda \|\beta^{(i)}\|_1 \\
  &=  \|p^{(i)} -  X \beta^{(i)}\|_{2}^{2} + \lambda \|\beta^{(i)}\|_1
\end{split}
\label{eq:loss}
\end{equation}
Here, $J$ is the error function and $\lambda$ is the hyper-parameter which controls the severity of the penalty on complex models. We do this for each of $n$ time series, so we will end up applying ordinary least squares $n$ times.

This model offers a sparse regression \cite{lasso} that selects only the time series that have predictive value and identifies the temporal correlation between these time series. The bias and size of the resulting model can be set by modifying the $L_1$ regularization parameter $\lambda$. This is crucial since the smaller the model is, the less coefficients it has, and thus, the less data it needs to make predictions. Since retrieving data over a network has latency, and low latency when detecting anomalies is a priority in this algorithm, the size of the data to be retrieved should be as small as possible without greatly affecting the model's accuracy.

Finally, we store the estimated standard error corresponding to each of the $n$ time series to be used later for anomaly detection:
\begin{equation*}
\sigma^{(i)}=\sqrt{\frac{\sum\limits_{t=w+1}^h \Big(\epsilon_t^{(i)}\Big)^2}{h - m}}, \text{ for } i = 1, ..., n.
\end{equation*}
Here, $m$ is the number of non-zero coefficients. This estimator is motivated by the unbiased estimator for the error variance in a simple linear regression without regularization. The performance of its can be seen in Reid, Tibshirani, and Friedman \cite{RTF2014}.  
 
\section{Detection Model for Real-Time Applications}
\label{sec:detecting}
The model described in Section \ref{sec:model} makes real time predictions of all time series in the system as it streams new data. In order to detect anomalies, we compared the prediction the model does at current time $now$, $p_{now}^{(i)}$, with its real value coming from the data stream, $s_{now}^{(i)}$ by running a t-test with the t-statistic defined in equation (\ref{eq:t_one_example}):
\begin{equation}
\label{eq:t_one_example}
    t^{(i)} = \frac{s_{now}^{(i)}-p_{now}^{(i)}}{\sigma^{(i)}}.
\end{equation}
This test gives us a p-value we use to compare against a given threshold p-value-threshold, which is a parameter of the model. If the p-value we get with t-statistic (\ref{eq:t_one_example}) is smaller than p-value-threshold, then we are in presence of an anomaly and an alert should be raised.

The parameter p-value-threshold is probably the most important parameter in the model because it directly specifies how sensitive the model is in detecting anomalies. Usually its value is very low. A value of around 1e-5 would roughly detect an anomaly every 100,000 minutes.

It's important to note that this Detection procedure has very low latency because the model is sparse and it only needs to make few calculations in the linear combination. This is a key feature that puts this model ahead of lower latency, but accurate, models like DPCA \cite{russell}.

There is another aspect of real life systems that has not been covered so far. Since in real life systems, random and short spikes can be normal and not anomalies, we use a smoothed anomaly definition as opposed to the naive comparison. Concretely, since these spikes may result in false positive classification of anomalies instead of testing for an anomaly with just the current data-point $s_{now}^{(i)}$, we test for an anomaly with a sample of data-points. This sample of size $d$ for time series $i$ is defined as
\begin{equation*}
    a^{(i)} = \left\{s_k^{(i)}: k \in [now-d+1,..., now]\right\}.
\end{equation*}
Our model checks the probability that $a^{(i)}$ comes from a normal distribution with mean $p_t^{(i)}$ by running a t-test with the t-statistic defined by
\begin{equation*}
    t^{(i)} = \cfrac{\cfrac{1}{d}\left(\sum\limits_{k=now-d+1}^{now} s_k^{(i)}\right)-p_{now}^{(i)}}{\cfrac{\sigma^{(i)}}{\sqrt{d}}}.
\end{equation*}
Here, $\displaystyle \frac{\sigma^{(i)}}{\sqrt{d}}$ is the sample $a^{(i)}$ standard error.

\section{Detection Model - Preliminary Results}
\subsection{Baseline}
\begin{figure}[h]
\begin{center}
\includegraphics[scale=0.36]{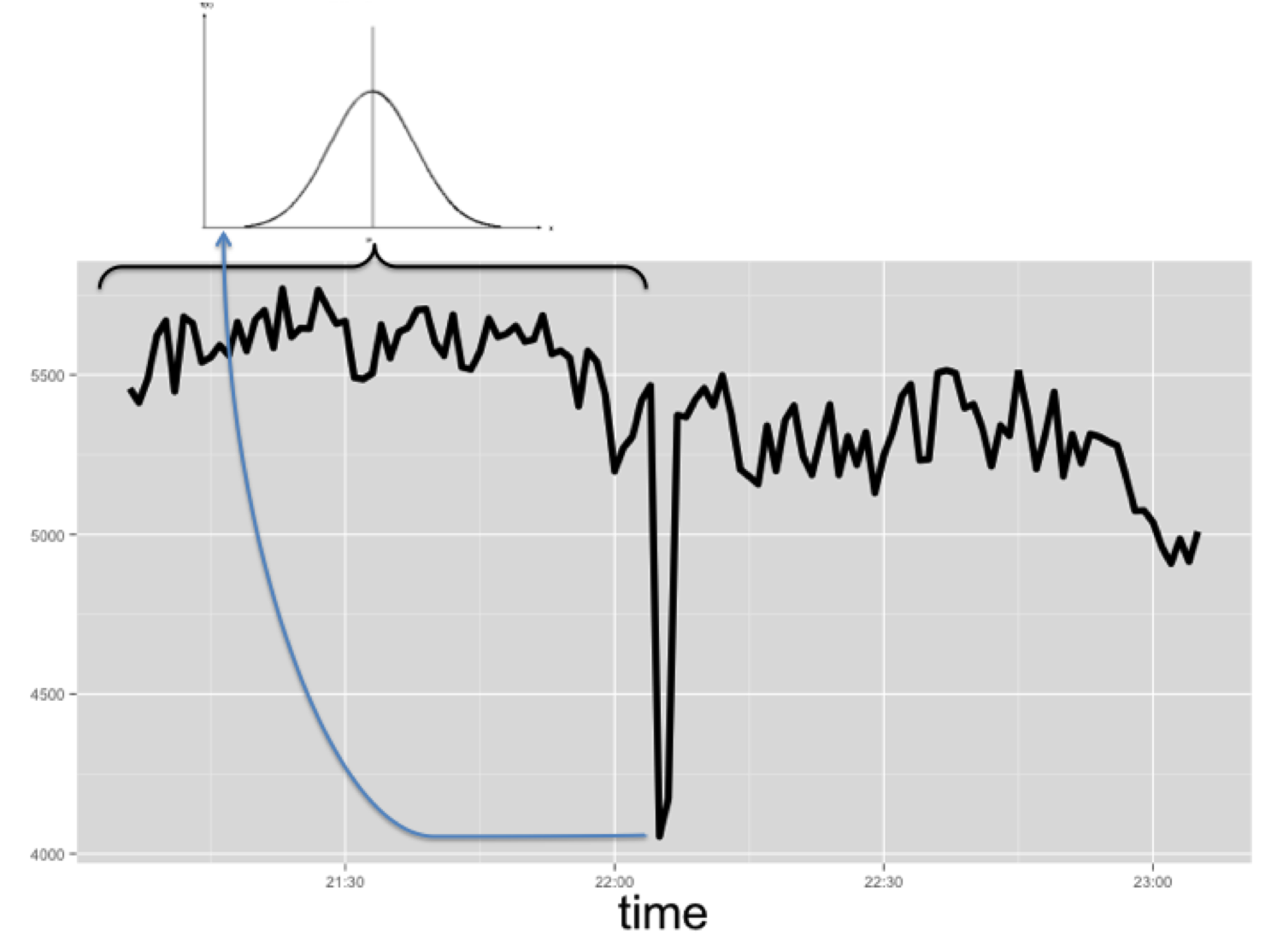}
\caption{\label{gaussian}The Gaussian Detector.}
\end{center}
\end{figure}

As a baseline to highlight the difficulty of the problem we are trying to solve, we implement and test a model that characterizes the time series as multivariate Gaussian. More concretely, we build a generative model where time series are generated from an $m$-dimensional multivariate Gaussian distribution and then set a threshold of the probability of a time series deviating from its mean. We can visualize this method in Figure \ref{gaussian}.

\subsection{Oracle}
Again, highlighting the difficulty of the problem that we are trying to solve, we present an oracle where we use the test set as the training set and calculate the ``test error," since we know that the test error should be bounded by the training error, this oracle will allow us to understand the difficulty of our problem definition. The results shown in section III, compare the error of the oracle formulation above.

\subsection{Performance Comparison}
We compare our model's performance with our baseline model  (multivariate Gaussian \cite{cs229}) and with the state of the art Dynamic Principal Components model (DPCA) \cite{russell}. We run the models on a year worth of server logs data, aggregated as $100,000$ time series. As noted in section \ref{sec:data}, the data are labeled by humans (with domain knowledge) as anomaly or not anomaly. We use an auto regression window size $w$ of one week. The output of each model by minute (anomaly or not anomaly class) is compared against the real label of the data. Finally, we calculate the $F1$ score to evaluate the performance of each model.

\begin{table}[h]
\begin{center}
\caption{Accuracy}\label{table:f1}
{
\begin{tabular}{@{}lr@{}}
  \hline \hline
Algorithm & $F_1$ Score \\ \hline
Gaussian Model (Baseline) & $0.33$ \\
DPCA                      & $0.92$ \\ 
Proposed Algorithm        & $0.90$ \\ \hline
\end{tabular}
}
\end{center}
\end{table}
Table \ref{table:f1} shows that even though our algorithm's accuracy is below DPCA's accuracy, our preliminary model is able to achieve relatively high performance. 

\begin{table}[h]
\begin{center}
\caption{Latency}\label{table:latency}
{
\begin{tabular}{@{}lr@{}}
  \hline \hline
Algorithm & latency (millsecs) \\ \hline
Gaussian Model (Baseline) & $40$ \\
DPCA                      & $2,432$ \\ 
Proposed Algorithm        & $12$ \\ \hline
\end{tabular}
}
\end{center}
\end{table}

Table \ref{table:latency} shows that our model is faster than the very accurate DPCA. It is also faster than the Gaussian model, which is explained by the fact that our model is sparse and does not need to load all the data in the auto-regression window, just the data points that are highly correlated.

Since low latency is a crucial component in real time applications, we believe our model's combination between good accuracy and extremely low latency is very powerful.

\section{Bayesian Network Approach For Characterizing Anomaly Likelihood}

\subsection{Nodes}
The second part of this piece deals with building a Bayesian Network to characterize the probability of an anomaly. Recall that for every time series $i$ we run a distributed lag regression (section \ref{sec:model}) where the regressors are lag variables of $i$ and every other time series $j \in \{1, ..., n\}$. The model for time series $i$ is parametrized by a vector $\beta^{(i)} \in \mathbb{R}^{wn}$ of coefficients on the regressors (the $L_1$ regularization term ensures that the majority of these weights are set to zero). 

We denote by $\beta'^{(i)}$ the vector of non-zero weights for time series $i$, and by $\beta'^{(i)}_{j,k}$ the non-zero weight component for time series $i$ that corresponds to lag regressors of time series $j$ with a lag of $k$ (as in Section \ref{sec:model})

Similarly, let $\gamma'^{(i)}_{j,k}$ correspond to the lag \textit{variables} with non-zero weights for time series $i$ regressing on lag regressors of time series $j$ where the lag for that regressor is $k$. Let the length of this vector of variables be $d_{i,j}$.

The random variables in our network include: $\widehat{\epsilon^{(i)}}$ for $i \in \{1, ..., n\}$; $A_i$ for $i \in \{1, ..., n\}$; and $\gamma'^{(i)}_{j,k}$ for $i \in \{1, ..., n\}$, $j \in \{1, ..., n\}$, and $k \in \{1, ..., d_{i,j}\}$.

Note that although there may be overlap between $\gamma'^{(i)}_{j}$ and $\gamma'^{(l)}_{j}$ for $i \ne l$, our Bayesian network considers the set of all these variables.

\subsection{Domain}
\begin{figure}[h]
 \tikzset{
   every neuron/.style={
     circle,
     draw,
     minimum size=0.75cm
   },
   neuron missing/.style={
     draw=none, 
     scale=1,
     text height=0.1cm,
     execute at begin node=\color{black}$\vdots$
   },
   neuron cont/.style={
     draw=none, 
     scale=1,
     text height=0.1cm,
     execute at begin node=\color{black}$\cdots$
   },
 }

 \begin{center}
 \begin{tikzpicture}[x=1cm, y=1cm, >=stealth]

  \foreach \m/\l [count=\y] in {1,2,3}
   \node [every neuron/.try, neuron \m/.try] (input-\m) at (0, 7 - 2 * \y) {$\gamma'^{(i)}_{\m, 1}$};
 
  \foreach \m/\l in {missing}
   \node [every neuron/.try, neuron \m/.try] (input-\m) at (0, 7 - 2 * 4) {};

 \foreach \m/\l in {4}
   \node [every neuron/.try, neuron \m/.try] (input-\m) at (0, 7 -  2 * 5) {$\gamma'^{(i)}_{n, 1}$};
   
  \foreach \m/\l [count=\y] in {1,2,3}
   \node [every neuron/.try, neuron \m/.try] (hidden1-\m) at (3, 7 - 2 * \y) {$\gamma'^{(i)}_{\m, 2}$};
 
  \foreach \m/\l in {missing}
   \node [every neuron/.try, neuron \m/.try] (hidden1-\m) at (3, 7 - 2 * 4) {};

 \foreach \m/\l in {4}
   \node [every neuron/.try, neuron \m/.try] (hidden1-\m) at (3, 7 -  2 * 5) {$\gamma'^{(i)}_{n, 2}$};
   

\foreach \m [count=\y] in {cont, cont, cont, missing, cont}
   \node [every neuron/.try, neuron \m/.try ] (hidden2-\m) at (5, 7 - \y*2) {};

  \foreach \m/\l [count=\y] in {1,2,3}
   \node [every neuron/.try, neuron \m/.try] (hidden3-\m) at (7, 7 - 2 * \y) {$\gamma'^{(i)}_{\m, d_{i, \m}}$};
 
  \foreach \m/\l in {missing}
   \node [every neuron/.try, neuron \m/.try] (hidden3-\m) at (7, 7 - 2 * 4) {};

 \foreach \m/\l in {4}
   \node [every neuron/.try, neuron \m/.try] (hidden3-\m) at (7, 7 -  2 * 5) {$\gamma'^{(i)}_{n, d_{i, n}}$};

 \foreach \m [count=\y] in {1}
   \node [fill=yellow, every neuron/.try, neuron \m/.try ] (output-\m) at (11, 7 - 2) {$\widehat{\epsilon^{(i)}}$};

 \foreach \m [count=\y] in {1}
   \node [fill=green, every neuron/.try, neuron \m/.try ] (anomaly-\m) at (13, 7 - 2) {$A_i$};

\draw [->] (output-1) -- (anomaly-1);

\foreach \i in {1,...,4}
 	\draw [->] (hidden3-\i) -- (output-1);

\foreach \i in {1,...,4}
	\draw [blue, ->] (hidden1-\i) to[out=-15,in=-90] (output-1);

\foreach \i in {1,...,4}
	\draw [red, ->] (input-\i) to[out=40,in=120] (output-1);
 \end{tikzpicture}
 \caption{\label{BayesNet}Bayesian Network Structure For Time Series $i$, for $i \in \{1, ..., n\}$.} 
 \end{center}
 \end{figure}
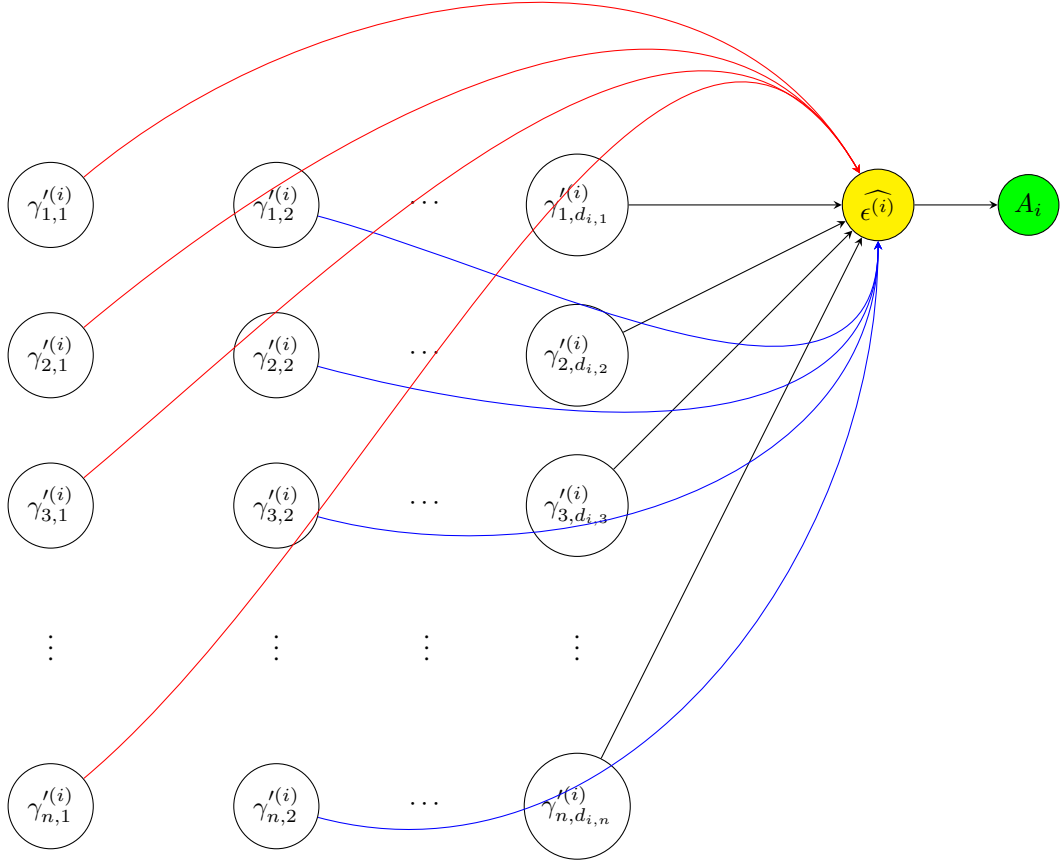

The lag variables $\gamma'^{(i)}_{j,k}$ represent the value of the lagged time series, their domain is continuous. The $A_{i}$ variables take values $\{0,1\}$ representing the presence of an anomaly at time $t+1$ and lastly the $\widehat{\epsilon^{(i)}}$ variables denote the how much our prediction deviates from the true value of the time series (See section 6.4). We can visualize the Bayes Net in Figure \ref{BayesNet}.  

\subsection{Factors}
The structure of the Bayesian Network is defined by adding, for each $i \in \{1, ..., n\}$, an edge from node $\gamma'^{(i)}_{j,k}$ to node $\widehat{\epsilon^{(i)}}$ for $j \in \{1, ..., n\}$, $k \in \{1, ..., d_{i,j}\}$, and an edge from $\widehat{\epsilon^{(i)}}$. 

\subsection{Theoretical Reasoning Underlying Bayesian Network Model}

Fan and Li \cite{iterated_lasso} show that under the approximation:
$$ \|\beta_j^{(i)}\|_1 \approx \frac{(\beta_j^{(i)})^2}{\|\beta_j^{(i)}\|_{1}}$$ 
The unconstrained formulation of the convex optimization problem as Equation (\ref{eq:loss}) can be solved by iterating:
\begin{equation*}
    \widehat{\beta^{(i)}} = \left(X^T X + \lambda \; (\text{diag} (|\beta^{(i)}|))^{-1}\right)^{-1} X^T p^{(i)}. 
\end{equation*}
Where $\widehat{\beta^{(i)}}$ is the estimate of $\beta^{(i)}$. 
Moreover, Fan and Li \cite{iterated_lasso} draw from the theory of k-step estimators in \cite{bickel} which proves that, with a good starting parameter, a on-step iteration can be as efficient as the penalized maximum likelihood estimator, when the Newton-Rapshon algorithm is used.

We have
\begin{equation*}
    \widehat{\epsilon^{(i)}} = p^{(i)} - X \widehat{\beta^{(i)}} = \left[I - X \left(X^T X + \lambda \; (\text{diag} (|\beta^{(i)}|))^{-1}\right)^{-1} X^T\right] \left(X \beta^{(i)} + \epsilon^{(i)}\right).
\end{equation*}
Thus, there is a direct relation between $\epsilon^{(i)}$ and $\widehat{\epsilon^{(i)}}$. As $\epsilon^{(i)}$ determines the possibility of anomaly, the derived relation gives rise to the appearance of the node $\widehat{\epsilon^{(i)}}$'s in the Bayesian network. 

\subsection{Challenges}

The generative model presented above proved to have many challenges. For concreteness, we highlight some of them.
  
Our initial attempt was discretizing the continuous random variables and using Maximum Likelihood method, an approach which is sometimes used in literature (\cite{PGM} p. 186). Note however that discretization provides a trade-off between accuracy of the approximation and cost of the computation (\cite{PGM} p. 607). Work done using this approach proved to be nonviable. For concreteness, note that the average number of parents ($\gamma'^{(i)}_{j,k}$'s) for each time series $i$ are $\sim75$. Even for a coarse discretization scheme of 10 bins, each local CPD table would have $\sim10^{75}$ entries, even with the decomposibility of factors in Bayesian Networks, the amount of data needed to accurately estimate the maximum likelihood (not to mention the computational complexity of collecting the sufficient statistics) would be intractable. Even if we could estimate the local maximum likelihood, the complexity of many inference algorithms is exponential in the size of the domain of the Conditional Probability Distributions (e.g. Variable Elimination). One possible approach that we are exploring is tuning the hyperparameter of the cost of regularization term to restrict the number of variables in each time series to be used in the Bayesian network. 
  
Another approach we are exploring is using a Hybrid Model, which contains both continuous and discrete random variables. This structure in and of itself presents many challenges (see, e.g., \cite{PGM}, chapter 14). In parcticular, we can show that inference on this class of networks is \textit{NP-Hard}, even when the network structure is a polytree (\cite{PGM} p. 615).

\subsection{Practical Approach}
Solving the full Bayesian network model is super complicated as discussed above. Thus, we solve a simplified version of this problem. Specifically, we ignore the prior distribution and use the conditional distribution as the posterior distribution. Also, the relation between $\widehat{\epsilon^{(i)}}$ and $A_i$ is linear. The detailed procedure for characterizing anomalies is provided below. We focus on detecting anomalies for one time series only; other time series in the Bayesian network should work in the similar way.

\begin{itemize}

    \item Step 1: We normalize $\widehat{\epsilon^{(i)}}$ to be in the $(0, 1)$ interval.

\begin{equation*}
    \widehat{\epsilon^{(i)}} \leftarrow \frac{\widehat{\epsilon^{(i)}} - min_\epsilon + 1}{max_\epsilon - min_\epsilon + 2},
\end{equation*}
where $min_\epsilon$ is the minimum over all observed values of $\widehat{\epsilon^{(i)}}$ and $\max_\epsilon$ is the maximum over those values. The constants $1$ and $2$ make the normalized $\widehat{\epsilon^{(i)}}$ fall strictly in $(0, 1)$.  
    
    \item Step 2: We specify the relation between $\widehat{\epsilon^{(i)}}$ and $\gamma'^{(i)}_{j,k}$'s as a linear Gaussian model. More precisely, we determine the relation
\begin{equation*}
    \widehat{\epsilon^{(i)}} = \sigma\left(\sum_{\substack{j = 1, ..., n \\ k = 1, ..., d_{i, j}}} \gamma_{j, k}^{\prime(i)} \cdot c_{j, k}\right) \text{ with coefficients } c_{j, k}'s \text{ where } \sigma(x) = \frac{1}{1 + e^{-x}}. 
\end{equation*}
This relation can be rewritten as
\begin{equation*}
    \log\left(\frac{\widehat{\epsilon^{(i)}}}{1 - \widehat{\epsilon^{(i)}}}\right) = \sum_{\substack{j = 1, ..., n \\ k = 1, ..., d_{i, j}}} \gamma_{j, k}^{\prime(i)} \cdot c_{j, k}, 
\end{equation*}
which can then be solved by least square method. 
    
    \item Step 3: We specify the probability that $A_i = 1$ given $\widehat{\epsilon^{(i)}}$ that
\begin{equation*}
    P\left(A_i = 1 | \widehat{\epsilon^{(i)}}\right) = \min\left(1, \alpha \times \widehat{\epsilon^{(i)}}\right),
\end{equation*}
then $A_i$ is characterized as $1$, which is anomaly, if and only if this probability is greater than $0.5$ and $0$ otherwise. The constant $\alpha$ can be chosen using cross-validation so as to maximize the $F_1$ score. 
        
\end{itemize}

\subsection{Result}
In this section we use four time series as our data set in order to characterize anomalies for one of these four time series. The reason we drastically reduce the size of the data is because matrix $X \in \mathbb{R}^{(h-w) \times wn}$ (as defined in equation (\ref{eq:loss})) and limiting the size of our data set was the only way to be able to compute this in a single machine. Our data set has the following size:

\begin{itemize}
    \item $h = 43,200$ (30 days of minute data)
    \item $w = 7,200$ (5 days of minute data)
    \item $n = 4$ (four time series)
\end{itemize}

With this data set we generate a matrix $X$ of float elements and dimensions $36,000$ rows by $28,800$ columns. This matrix has a size of roughly 17GB (depending on the platform used).

Finally, we obtain $\alpha = 5/6$ and the corresponding $F_1$ score is $0.78$. 

\subsection{Error Analysis}
This value is certainly not as good as that of DPCA or our original discriminative anomaly detection approach, but the comparison is nuanced since above we only processed one time series with a simplified Bayesian Network (because of run time complexity). Notwithstanding, this result is expected; discriminative models will typically outperform generative models when the relationships expressed by the generative model only approximates the true underlying generative process (particularly for our case where we follow some approximations to simplify the model complexity) in terms of classification error rate.

However, the advantage of our novel approach is that our model is richer. As a generative model, we can make explicit claims about the process that underlies the time series. Future work may included running inference queries on the generative model (after tuning it for better performance) to better understand the underlying process where the data is coming from.

\section{Conclusion and Future Work}
Equation (\ref{eq:loss}) proposes a loss function $J$ composed by an $L_2$ error and an $L_1$ regularization. We think it is worth to test if an $L_1$ error and an $L_1$ regularization performs better. The reason behind this is that an $L_1$ error is more biased against outliers. If our regression $p^{(i)}$ is less affected by the anomalies in the training data set, then it will predict anomalies more accurately in the testing data.

Also, the model proposed in section \ref{sec:model} may show false positives when the predictors have an anomaly. This is because after an anomaly exists, it still stays in the data that will be later use as an input in out regression. We did not see this in our experiment, but it is something we would like to test and evaluate.

\end{document}